\documentclass[10pt]{article}
\usepackage{mathptmx}
\usepackage[margin=.5in]{geometry}
\usepackage{wrapfig}
\usepackage[export]{adjustbox}
\usepackage{subcaption}
\usepackage[utf8]{inputenc} 
\usepackage[T1]{fontenc}    
\usepackage{hyperref}       
\usepackage{url}            
\usepackage{booktabs}       
\usepackage{amsfonts}       
\usepackage{nicefrac}       
\usepackage{microtype}      
\usepackage{graphicx}

\begin{document}

\begin{center}
  Mason R.~Swofford \\
  Department of Computer Science\\
  Stanford University\\
  \texttt{mswoff@stanford.edu} \\ 
  https://github.com/mswoff/Image-Completion-on-CIFAR-10\\
  \LARGE Image Completion on CIFAR-10
\end{center}


\begin{abstract}
This project performed image completion on CIFAR-10, a dataset of 60,000 32x32 RGB images, using three different neural network architectures: fully convolutional networks, convolutional networks with fully connected layers, and encoder-decoder convolutional networks. The highest performing model was a deep fully convolutional network, which was able to achieve a mean squared error of .015 when comparing the original image pixel values with the predicted pixel values. As well, this network was able to output in-painted images which appeared real to the human eye.
\end{abstract}

\section{Introduction}	
Image completion, or image in-painting, is a fascinating and quickly growing area of research in machine learning. Given an image with some pixels removed, the image completion task is to correctly fill in those missing pixels. Therefore, the input to an image completion algorithm is an image with some pixels removed, and the algorithm outputs its estimate of the values of those missing pixels. This has various use cases. For instance, if an image is partially corrupted, the corrupted pixels can be removed, and an image completion algorithm can be used to fill in the missing pixels so that the image looks like the original, uncorrupted version. As well, object removal can be accomplished with image completion. The pixels corresponding to the object can be removed, and then an image completion algorithm can fill in the missing pixels values in accordance with the rest of the image.\\

This can be a challenging problem, because in some cases a large portion of the image is removed, and the algorithm must try and recover the missing information using only the remaining parts of the image. In fact, this is not a problem most humans perform well at, as it takes artistic abilities to make an image look natural. My problem will face further challenges, as I chose to use very small images which therefore contain less information in them. My approach was to remove a square from the center of each image using a mask and then apply a variety of different convolutional networks to the masked images, to see which could best predict the original images. I drew inspiration for my network architectures from some of the networks presented in the next section.

\section{Related work}
Current state of the art in image completion largely relies on GANs (Generative Adversarial Networks), which are able to learn a distribution for images, and then try to fill in the missing pixel values according to this learned distribution. These networks are able to perform incredibly well even when a large portion of the original image is removed. However, GANs, and in particular DCGANs (Deep Convolutional Generative Adversarial Networks), require huge amounts of data, are typically only trained and used on one class of images (e.g. faces, scenery, or cats), and are employed on larger images which contain more pixels and therefore more information [2]. \\

Prior to DCGANs appearance, most image completion algorithms did not rely on machine learning, and instead focused on mathematical methods for in-painting. One interesting algorithm performed in-painting by scouring a large corpus of images to find semantically similar ones, and then in-painted the corrupted image by taking fragments from these similar corpus images [5]. Other techniques tried to find similar patches within the same image that could fill in the missing pixels while satisfying certain constraints [4].\\

In researching prior work on image completion, I noticed parallels between image completion and image segmentation, where the input is an image and the output is a map of which pixels correspond to different objects in the image. Both of these tasks require high dimensional outputs, which is different from the most common computer vision task, classification. As well, if an algorithm can identify which pixels correspond to an object, then I theorized if some of those pixels are removed, it can more easily in-paint them by using information from the object. These ideas lead me to consider networks used in image segmentation. \\

The Fully Convolutional Network (FCN) used by Long and Shelhamer was one of the first networks to perform exceptionally well on image segmentation. Unlike previous work, their network only used convolutional layers and could therefore be used on inputs of arbitrary sizes [7]. The simplicity of their approach along with the high performance motivated me to try this model as my initial approach.\\

SegNet built on FCN by employing an encoder-decoder architecture where the image was first transformed into a low dimensional encoding, and then that low dimensional encoding was up-sampled to produce an output of the same size as the input [3]. The idea of using encoders and decoders has yielded astounding results in fields outside of image segmentation as well, most notably in machine translation.

\section{Dataset and Features}
\begin{wrapfigure}{r}{0.25\textwidth}
    \centering
    
    \includegraphics[width=0.25\textwidth]{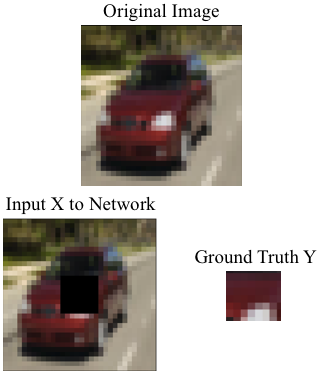}
    \caption{\small Sample Training Example}
\end{wrapfigure}
As mentioned, I wanted to see how image completion algorithms would fare on smaller datasets with both fewer training examples and smaller image sizes. As well, I wanted the dataset to include multiple object classes, to see how well an algorithm could perform under tougher conditions. I settled on the CIFAR-10 dataset because it met all of these conditions. CIFAR-10 is a subset of the Tiny Images dataset which contains 80 million 32x32 RGB images. CIFAR-10 includes just 10 object classes, with 6,000 images per class for a total of 60,000 32x32 RGB images. These classes vary widely, from airplanes to frogs to trucks [6]. The data was split into 50,000 train images, 5,000 development images, and 5,000 test images. Since CIFAR-10 was intended for object classification, some pre-processing was required to transform the dataset.\\

Instead of feeding each image to the network directly, a mask was applied that zeroed out the center 8x8 pixels. This 8x8 region was now what the network was trying to predict. Therefore, the true label for each image was the inverse mask applied to the image, resulting in an 8x8 RGB label. This label was flattened into a 192 (8x8x3) dimensional vector, so that it could be more easily fed into a loss function. The pixel values were also scaled from the usual [0,255] range down to the [0,1] range. An example of this process is shown in Figure 1.

\section{ Methods }
In total, 6 different networks were trained and compared. While there were 6 total networks, they fall into just 3 different categories: fully convolutional network, convolutional network with fully connected layers, and encoder-decoder convolutional network. Regardless of the algorithm used, each network was given a masked-out image and had to predict the 192-dimensional output. Mean squared error was used as the loss function for all of the algorithms. It computes the mean squared difference between each element in the predicted 192-vector and the label 192-vector. This means that it compares how close each channel value (red, green, and blue) is in the prediction versus the ground truth. By optimizing mean squared error, the algorithms learn to match the prediction to the true value for each pixel. The formula for mean squared error is as follows:
$$MSE = \frac{1}{192} \sum_{i=1}^{192}(y_i - \hat{y}_i)^2 $$ 

As well, all the networks were trained using mini-batch gradient descent with the Adam optimization algorithm with variable learning rates which were hand tuned. All layers (except max pooling and flattening layers) use rectified linear activations. Since the final per channel output should be in the range [0,1], sigmoid activations were used at first. However, they proved to work poorly and instead the rectified linear activation was used even for the final layer, and it was clipped to be in the range [0,1].\\

3 fully convolutional networks were trained: a shallow, deep, and super deep network. By only including convolutional and max pooling layers and no deconvolutions or fully connected layers, these networks did not have access to the same amount of global information as the other networks and were theorized to over-fit the training data less. The deeper networks not only have more layers, but they also have more filters per layer. This allows them to learn more features from the input image (e.g. texture patterns, lines, etc.). The kernels used along with the input size at each layer are as follows, where the part in brackets is the kernel applied and number of filters used is in parentheses. For example, 32x32x3    ->[5x5 Convolution (10)]->    28x28x10 means that 10 5x5 filters were applied to the 32x32x3 input and resulted in a 28x28x10 output for that layer. Note: all convolutions are “valid” convolutions unless otherwise noted, and all max pooling layers are 2x2 filters with strides of 2.\\

Shallow Network: 32x32x3    ->[5x5 Convolution (10)]->    28x28x10    ->[5x5 Convolution (20)]->    24x24x20    ->[Max Pooling]->    12x12x20    ->[5x5 Convolution (20)]->    8x8x20    ->[1x1 Convolution (3)]->    8x8x3    ->[Flatten]->    192x1\\

Deep Network: 32x32x3    ->[5x5 Convolution (20)]->     28x28x20    ->[5x5 Convolution (40)]->      24x24x40    ->[Max Pooling]->    12x12x40    ->[3x3 Convolution (60)]->    10x10x60    [3x3 Convolution (80)]->    8x8x80    ->[1x1 Convolution (3)]->    8x8x3    ->[Flatten]->    192x1\\

Super Deep Network: 32x32x3    ->[7x7 “Same” Convolution (40)]->    32x32x40    ->[7x7 “Same” Convolution (40)]->    32x32x40    ->[7x7 Convolution (40)]->    26x26x40    ->[7x7 Convolution (60)]->     20x20x60    ->[5x5 Convolution (60)]->    16x16x60    ->[Max Pooling]->    8x8x60    ->[5x5 Convolution (60)]->    4x4x60    ->[4x4 Convolution (192)]->    1x192\\

A convolutional network with fully connected layers was also trained. This model was based on the models typically used in object classification, where first convolutions are applied, and then the output is flattened and fed into a series of fully connected layers and finally a softmax layer. However, my network differed from this in that I did not have a softmax layer, and instead of having just one output, I had 192. Following the previous conventions, the network is as follows.\\

Fully Connected Network: 32x32x3    ->[5x5 Convolution (10)]->    28x28x10    ->[5x5 Convolution (20)]->    24x24x20    ->[Max Pooling]-> 12x12x20    ->[5x5 “Same” Convolution (30)]->    12x12x30    ->[3x3 “Same” Convolution (40)]->    12x12x40    ->[Flatten]->    5,760x1    ->[Fully Connected Layer]->    768x1    ->[Fully Connected Layer]-> 192x1\\

The last models tried were 2 encoder-decoder networks. These models first created a lower dimensional representation of the missing pixels, and then use deconvolutions to up-sample the representation and create the final output. By first creating a lower dimensional encoding of the missing pixels, encoder-decoder networks tend to over-fit less, as they can only store so much information in the lower dimension. The deconvolutions used were all 3x3 kernels with strides of 2. Following the previous conventions, the networks are as follows.\\

Encoder Network: 32x32x3    ->[5x5 Convolution (10)]->    28x28x10    ->[5x5 Convolution (20)]-> 24x24x20    ->[Max Pooling]->    12x12x20    ->[5x5 Convolution (30)]->    8x8x30    ->[5x5 Convolution (128)->    4x4x128    ->[3x3 Deconvolution (3)]->    6x6x3    ->[3x3 Deconvolution (3)]->    8x8x3    ->[Flatten]->    192x1\\

Deep Encoder Network: 32x32x3    ->[5x5 Convolution (30)]->    28x28x30    ->[5x5 Convolution (40)]->    24x24x40    ->[Max Pooling]->    12x12x40    ->[5x5 Convolution (40)]->    8x8x40    ->->[5x5 Convolution (40)]->    4x4x40    ->[Flatten]->    1x640    ->[Fully Connected Layer]->    1x768 ->[3x3 Deconvolution (40)]->    2x2x40    ->[3x3 Deconvolution (40)]->    4x4x40    ->[3x3 Deconvolution (3)]->    8x8x3    ->[Flatten]->    192x1\\

\section{Experiments/Results/Discussion}

Various hyper-parameters were experimented with over the course of this project, including the different model architectures, activation functions, and learning rates. I began with the idea to try at least one of each type of architecture: fully convolutional network, convolutional network with fully connected layer, and encoder-decoder network.\\

For the learning rate, I experimented with different values ranging from .005 to .00001. However, this was all done by looking at the development error and hand tuning the learning rate for each model. As well, I decreased the learning rate as the models trained.\\

The first model tested was the shallow fully convolutional network (Shallow Network). With this network, I tested the sigmoid verses rectified linear activation for the final layer, and quickly saw that the sigmoid had poor mean squared error compared to the rectified linear activation. So, for all subsequent models, I only used the rectified linear activation (clipped to be in the range [0,1]); although it is possible the sigmoid activation would have worked better for some of these networks, it would have doubled the number of models I had to test. Using the rectified linear activation, the algorithm achieved a train and development loss that were approximately equal, .0153 and .0154, respectively. Figure 2a shows the epoch vs loss graph for this model (the first 5 epochs are removed for all graphs). While I did not know what Baye’s error would be for this task, after analyzing the images output by the network, they were not as good as I knew was possible. This lead me to believe that the algorithm had a bias problem and lead me to my second model.\\
\begin{figure}[h]
	\vspace{-.25in}
	\centering
	\begin{subfigure}{.33\textwidth}
      	\includegraphics[width=\textwidth]{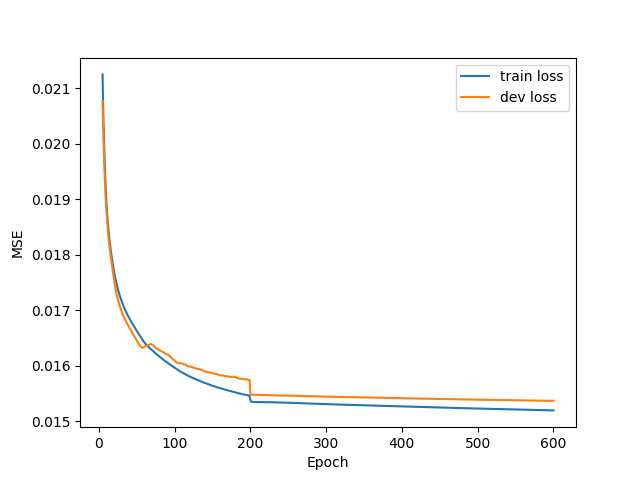}
      	\caption{\small Shallow Network}
	\end{subfigure}
    \begin{subfigure}{.33\textwidth}
        \includegraphics[width=\textwidth]{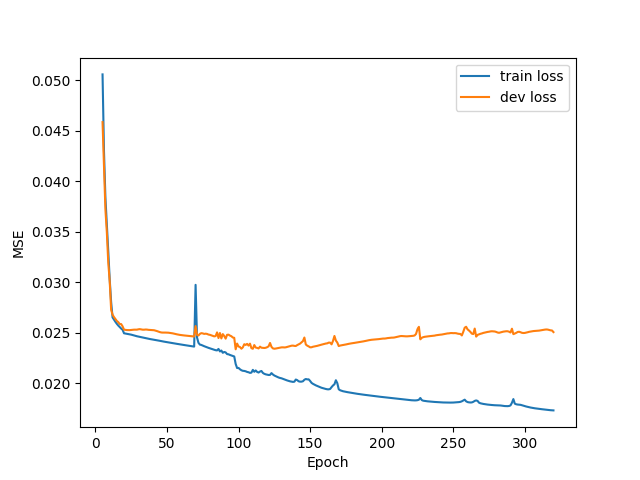}
    	\caption{\small Super Deep Network}
    \end{subfigure}
    \begin{subfigure}{.33\textwidth}
    	\includegraphics[width=\textwidth]{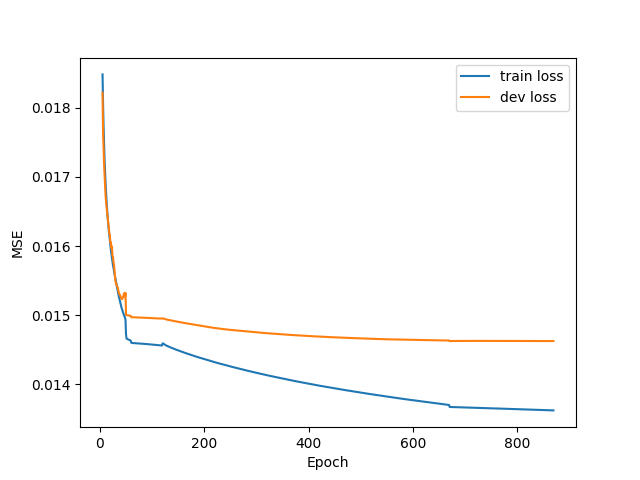}
    	\caption{\small Deep Network}
    \end{subfigure}
    \caption{\small Epoch vs Loss Graphs}
    \vspace{-.1in}
\end{figure}

Super Deep Network had orders of magnitude more parameters due to having more layers and more filters per layer. This allowed the model to learn more complex features than Shallow Network. However, the network took up to a minute to run per epoch, compared to around 5 seconds for Shallow Network. As well, it became apparent that there was not enough data in the training set to feed this deeper network, as it was able to over-fit the training set with a train loss of .0161 versus a development loss of .0253 after 320 epochs. Figure 2b shows the epoch vs loss graph for this model. While the train loss would likely have continued to decline if I trained for more epochs, it was clear the model was overfitting the data and the development loss was increasing every epoch. I believe that more data likely would have reduced a lot of the variance, and I considered adding the CIFAR-100 dataset (60,000 additional tiny images) to the training set or performing data augmentation on CIFAR-10. However, since part of my goal was to see how these algorithms performed on limited data, I decided against it and instead tried a new network, which was somewhere between Super Deep Network and Shallow Network in complexity.\\

The next model, Deep Network, had 5 convolutional layers unlike Shallow Network’s 4 and Super Deep Network’s 7. As well, it had more filters than Shallow Network, which I believed to be the biggest reason Shallow Network could not fit the data. Deep Network was able to perform much better than either previous network, achieving train and development losses of 0.013620 and 0.014624, respectively, after 870 epochs. Figure 2c shows the epoch vs loss graph for Deep Network.\\

Next, I built a model similar to Shallow Network but with 2 fully connected layers at the end. This model, Fully Connected Network, was able to fit the training set exceptionally well. Figure 3a shows the epoch vs loss graph for this model. However, it was unable to achieve development loss as low as Deep Network. The network clearly had a huge variance problem and lead me to abandon exploring more fully connected networks. I believed the issue was that predicting a high dimensional output using fully connected layers was too ripe for over-fitting.\\

The final class of models, the encoder-decoder networks, both performed reasonably well. Encoder Network achieved minimum development loss after 440 epochs, with a train and development loss of 0.015688 and 0.016934, respectively. Figure 3b shows the epoch vs loss graph for this model. After training for another 200 epochs, the algorithms development loss increased slightly and its training loss decreased slightly; however, it appeared to have almost leveled off, signifying that the model likely was under-fitting the data.\\

The final model, Deep Encoder Network, however over-fit the data. It achieved a training loss as low as .010 and development loss of .018 after 440 epochs. Figure 3c shows the epoch vs loss graph for this model. This network utilized a fully connected layer to help make the encoding, which could have contributed to it overfitting the training data. \\

\begin{figure}[h]
	\vspace{-.25in}
    \centering
    \begin{subfigure}{.33\textwidth}
      \includegraphics[width=\textwidth]{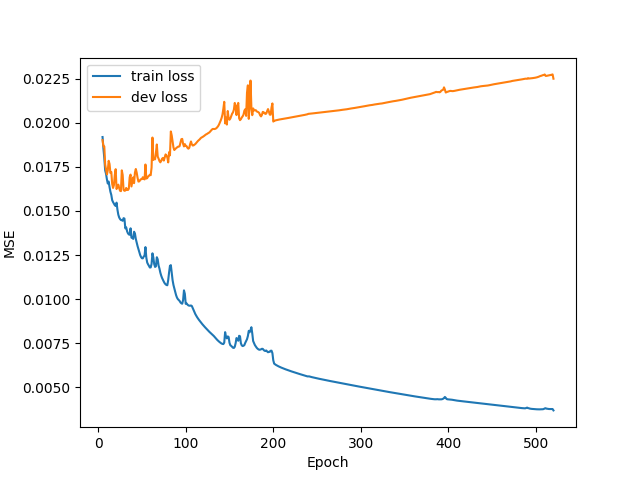}
      \caption{\small Fully Connected Network}
    \end{subfigure}
    \begin{subfigure}{.33\textwidth}
      \includegraphics[width=\textwidth]{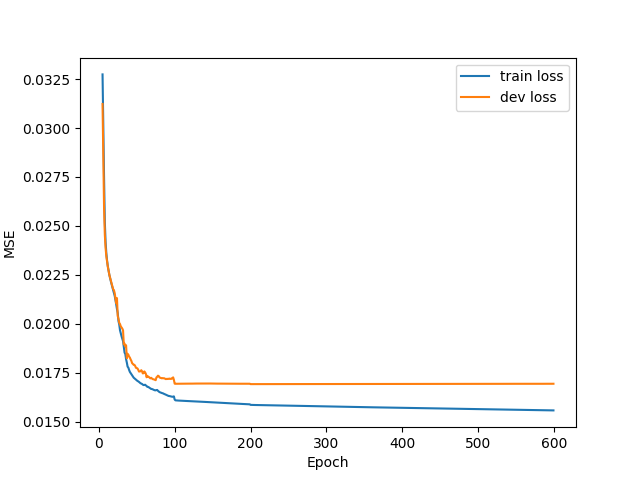}
      \caption{\small Encoder Network}
    \end{subfigure}
    \begin{subfigure}{.33\textwidth}
      \includegraphics[width=\textwidth]{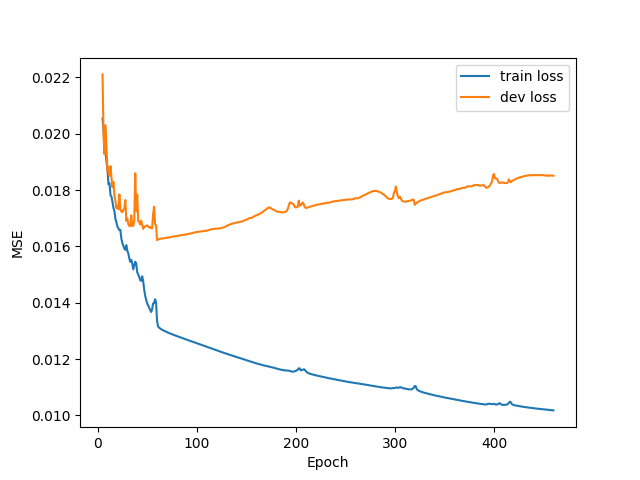}
      \caption{\small Deep Encoder Network}
    \end{subfigure}
    \caption{\small Epoch vs Loss Graphs}
    \vspace{-.1in}
\end{figure}

In order to better understand the mistakes the algorithms were making, I analyzed some of the training and development images output by the networks (Note: the algorithms only output the center square and I recombined this with the input image). In Figure 4, I compared the output of Deep Network versus the output of Fully Connected Network on two sample development set images. While I used means squared error as my loss function, my end goal was an algorithm that could fill in the missing pixels to create realistic looking images. Therefore, here I was looking to see whether a human could tell that the algorithms' outputs were not real images. In particular, I was trying to see if it was obvious that the 8x8 center region had been altered. Looking at 4b, this image appeared realistic enough to be an unmodified image. However, in 4c, the center 8x8 region clearly had been altered. In comparing 4b and 4c with the original image, 4a, it was easier to see that both images had been modified. However, the Deep Network's output was still pretty good.\\

One common area where all the algorithms failed was images with sharp edges. Looking at 4e and 4f, both algorithms blurred the edge between the truck's cab and trailer, while the original image 4d had a sharp line here. It is possible that adding more filters to the initial layers of the networks would allow the algorithms to better identify edges; however, this would likely have lead to the over-fitting problems that plagued the deeper networks. This also showcases how mean squared error can be a bad metric at times. If the algorithms had output a straight line of red pixels on one side and white pixels on the other, but shifted the line to the left or white of where it was originally, the algorithm would have incurred a large MSE, but the image would have looked realistic. Instead, by blurring the line somewhat, it ensures that the pixel values are not too far off what they should be, and thus incurs a lower MSE on average. I looked into alternative loss functions to mitigate this issue, but did not find one which I felt worked as well as MSE.\\

\begin{figure}[h]
	\vspace{-.2in}
    \centering
    \begin{subfigure}{.2\textwidth}
      \includegraphics[width=\textwidth]{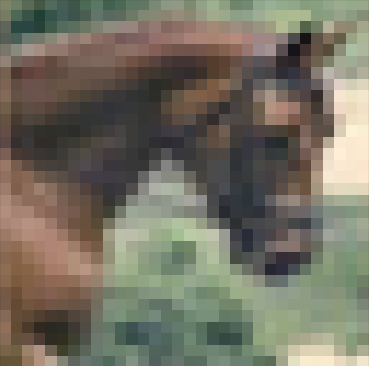}
      \caption{\small Original Image}
    \end{subfigure}
    \begin{subfigure}{.2\textwidth}
      \includegraphics[width=\textwidth]{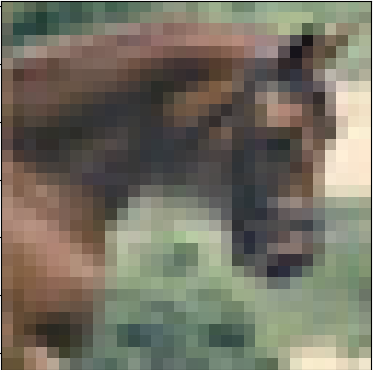}
      \caption{\small Deep Network}
    \end{subfigure}
    \begin{subfigure}{.2\textwidth}
      \includegraphics[width=\textwidth]{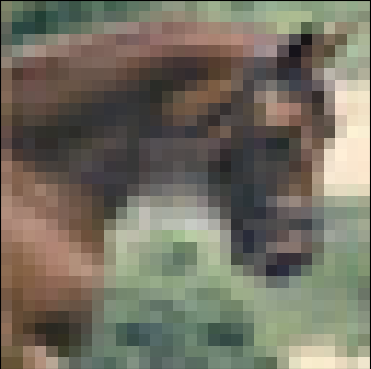}
      \caption{\small Fully Connected Network}
    \end{subfigure}
    
    \begin{subfigure}{.2\textwidth}
      \includegraphics[width=\textwidth]{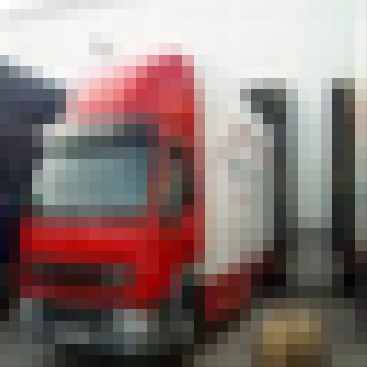}
      \caption{\small Original Image}
    \end{subfigure}
    \begin{subfigure}{.2\textwidth}
      \includegraphics[width=\textwidth]{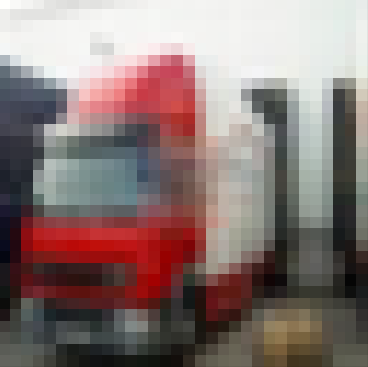}
      \caption{\small Deep Network}
    \end{subfigure}
    \begin{subfigure}{.2\textwidth}
      \includegraphics[width=\textwidth]{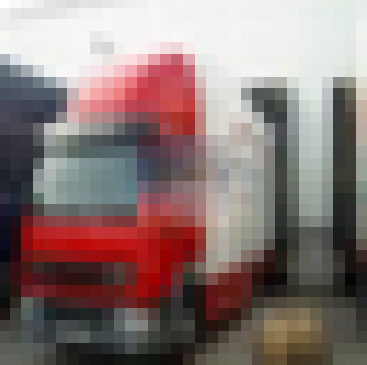}
      \caption{\small Fully Connected Network}
    \end{subfigure}
    \caption{\small Sample Network Outputs}
    \vspace{-.2in}
\end{figure}

\section{Conclusion/Future Work }
The goal of this project was to determine whether machine learning image completion algorithms could be trained on small images and small datasets and perform well. Figure 5 shows the MSE for the different algorithms on the test set, where the models are tested at their lowest development loss. From this chart and the development errors listed previously, Deep Network was the best performing model, with a test MSE of .015.\\
\begin{wrapfigure}{r}{0.25\textwidth}
    \centering
    
    \includegraphics[width=0.25\textwidth]{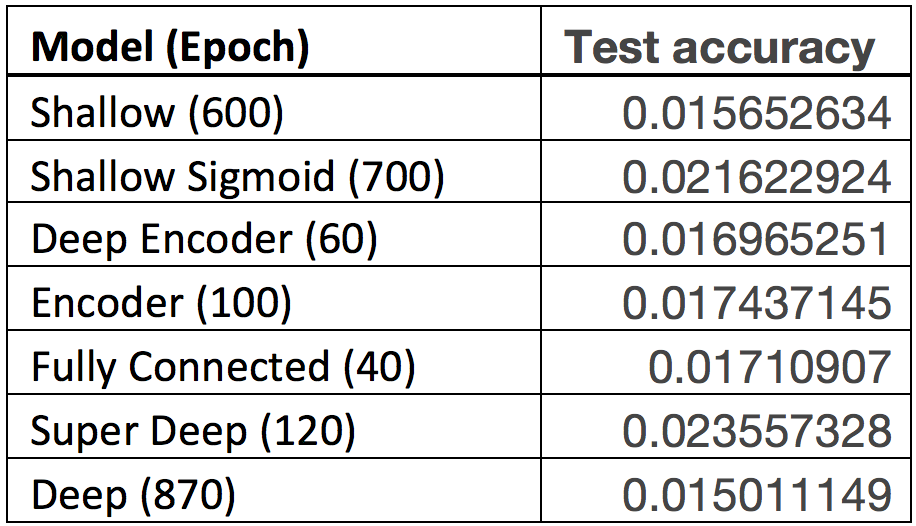}
    \caption{\small Test Losses}
\end{wrapfigure}

After analyzing some of the test images output by Deep Network, many of them looked realistic. However, images with certain defining features were almost never predicted correctly (as shown in Figure 4). This problem plagued all the networks and is one of the key benefits of using GANs. For instance, if an image of a human face with the eyes removed was fed to Deep Network, it would likely fill in the missing pixels with normal skin, as it performs a type of weighted average of the pixels surrounding the removed values. However, a GAN may have learned the distribution of human faces and recognize that it needs eyes in those locations to look realistic. This issue is in part due to my networks being trained on multiple classes of images, while GANs are often only trained on one class.\\

While the Deep Network performed the best on this dataset, if I increased the amount of training data, the deeper, more complex models would have likely outperformed it. This highlights one of the key issues in deep learning: you often need more data if you want to train a deeper model.\\

Given more time, I would like to experiment with this theory and try my models on a larger dataset with more images and possibly larger images. ImageNet's 64x64 RGB dataset would be perfect for this task and I think the deeper models would perform very well. Using larger datasets, I would likely be able to have more filters without risking over-fitting my training set. These additional filters could hopefully learn more complex features such as sharp edges, so that the algorithms would not make the same kinds of mistakes as they did on CIFAR-10.\\

As well, I would like to experiment more with encoder-decoder networks. I was only able to experiment with two different sized encodings, but I would like to see how larger and small dimensional encoding affect performance. I theorize that smaller encodings would reduce over-fitting and lead to interesting results. As well, it would be interesting to build a model that included encodings and had a skip connection with the input image, so that the model would have an easier time maintaining important features like sharp lines. 


\section*{References}
\medskip
\small
[1] M. Abadi, A. Agarwal, P. Barham, E. Brevdo, Z. Chen,
C. Citro, G. S. Corrado, A. Davis, J. Dean, M. Devin, S. Ghemawat,
I. Goodfellow, A. Harp, G. Irving, M. Isard, Y. Jia,
R. Jozefowicz, L. Kaiser, M. Kudlur, J. Levenberg, D. Mane,
R. Monga, S. Moore, D. Murray, C. Olah, M. Schuster,
J. Shlens, B. Steiner, I. Sutskever, K. Talwar, P. Tucker,
V. Vanhoucke, V. Vasudevan, F. Viegas, O. Vinyals, P. Warden, M. Wattenberg, M. Wicke, Y. Yu, and X. Zheng. 2015. TensorFlow: Large-scale machine learning on heterogeneous systems. www.tensorflow.org.\\

[2] Amos, B. 2018. Image Completion with Deep Learning in TensorFlow. bamos.github.io/2016/08/09/deep-completion/.\\

[3] V. Badrinarayanan, A. Kendall and R. Cipolla. 2017. SegNet: A Deep Convolutional Encoder-Decoder Architecture for Image Segmentation. IEEE Transactions on Pattern Analysis and Machine Intelligence, vol. 39, no. 12, pp. 2481-2495, Dec. 1 2017. doi: 10.1109/TPAMI.2016.2644615.\\

[4]  C. Barnes, E. Shechtman, A. Finkelstein, and D. Goldman. 2009.
Patchmatch: a randomized correspondence algorithm for
structural image editing. TOG, 2009.\\

[5] Hays, J., and Efros, A. A. 2007. Scene completion using
millions of photographs. ACM Trans. Graph (SIGGRAPH 2007)
26, 3.\\

[6] A. Krizhevsky. 2010. Convolutional Deep Belief Networks
on CIFAR-10. www.cs.toronto.edu/~kriz/cifar.html.\\

[7] Long, J., Shelhamer, E., and Darrell, T. 2014. Fully convolutional networks for semantic segmentation. CoRR,
abs/1411.4038.\\

\end{document}